# Deep Reinforcement Learning for Phishing Detection with Transformer-Based Semantic Features

Aseer Al Faisal

**ABSTRACT**

Phishing is a form of cybercrime in which people are deceived into exposing their personal information which can result in financial loss. These attacks are often executed via fraudulent messages, misleading ads, and breach of legitimate sites etc. This study uses Quantile Regression Deep Q-Network (QR-DQN) that combines RoBERTa semantic embeddings and crafted lexical features to enhance phishing detection that is aware of uncertainties. Instead of predicting mean returns, QR-DQN uses quantile regression to model the distribution over returns which improves stability and generalization for unknown phishing data over traditional RL DQN approaches to evaluate single scalar Q-values. A custom crawled diverse dataset of 105k URL was curated from PhishTank, OpenPhish, Cloudflare etc. The framework uses an 80/20 split of the dataset. The QR-DQN model with RoBERTa embeddings and lexical features achieved test accuracy 99.86%, precision 99.75%, recall 99.96% and F1-score 99.85% demonstrating high effectiveness. Compared to the standard DQN with lexical features, the suggested QR-DQN framework with lexical and semantic features lowers the generalization gap from 1.66 to 0.04 percent indicating great improvement in model generalization. Cross-validation experiments using 5-fold validation confirm the model robustness with mean accuracy of 99.90% and standard deviation of 0.04%. This shows the hybrid technique which combines quantile-based value estimate with RoBERTa semantic embeddings and lexical features is effective in recognizing serious phishing indicators and can adapt to evolving tactics while producing generalization tactics.

**Index Words**–Phishing detection, Deep reinforcement learning, RoBERTa semantic embeddings, Quantile Regression Deep Q-Network, QR-DQN, URL classification, lexical features, cybersecurity

## 1. INTRODUCTION

The modern internet faces persistent danger from evolving phishing threats consistently identified as one of the most critical attack vectors. What used to be a straightforward trick has quickly transformed into a complex, multi-faceted use issue. Phishing is a common cyber threat that coerces users into revealing sensitive information, including credentials and financial details. The cybercrime information center nearly detected 1M phishing incidents between November 2023 and January 2024 [1]. Phishing is executed by cyber criminals who possess an understanding of human psychology, which is mostly influenced by greed, gullibility, and the desire for exploration. The manipulation technique that lures humans into disclosing their private and sensitive info is known as social engineering. People's tendency to make mistakes or to trust them easily makes them a prime target for cyber threats that result in breaches of the security system. It is one of the most common techniques to manipulate the user into revealing their confidential information. The application of various forms of machine learning algorithms to detect phishing classification problems and in particular to security and malware detection has gained a lot of traction from the research community in recent years. These attacks intentionally trigger the victim psychologically and urges them for immediate action, manipulating trust, exploiting the tendency to comply with authority. This approach is really effective with social engineering attacks representing a significant portion of security breaches [2]. Even though one of the main underlying causes of Uniform Resource Locator (URL) is identifying theft and financial fraud through hijacking, ad injection attacks and URL spoofing [3].

The dynamic nature of these attacks where the defense signatures that were effective yesterday become predictable vulnerabilities today and has revealed a significant weakness in traditional static cybersecurity approaches. Most standard detection algorithms depend on the extraction of superficial characteristics referred to as engineered lexical features. These characteristics encompass metrics like the length of the URL, the count of special characters and the existence of particular keywords. This approach makes features weak against small changes by attackers that lead to big problems in generalization. This manifests as a challenge and a significant difference is often seen in controlled training and its ability to sustain that accuracy against unseen data and a variety of attack patterns. In the field of machine learning this limitation is called the train-test generalization gap ($G_{gap}$) where a large gap signifies that the detection policy has memorized the specific pattern of known attack instances in the training data rather than learning the transferable concepts of phishing URLs. Deep reinforcement learning (DRL) has emerged as a paradigm for tackling cybersecurity challenges especially via algorithms like DQN, DDQN which allow the agent to learn adaptive policies [4], [5]. Existing approaches show limitations in unpredictability and generalization. Unlike prior RL approaches relying exclusively on engineered lexical features and content based representations, our work advances by introducing Quantile Regression Deep Q-Network (QR-DQN) for distributional reinforcement learning. This hybrid approach combined with distributional value learning achieves a better generalization compared to standard DQN on identical features through uncertainty-aware, semantically-informed policy learning. This framework combines frozen RoBERTa transformer embeddings 768-dimensional semantic representations with lexical features which allows the agent to capture richer semantic interpretation of malicious intent and stronger generalization beyond training distributions such as domain impersonation, path obfuscation, brand mimicry that transcend superficial pattern matching.

Previous studies investigating Deep Reinforcement Learning (DRL) for URL state representations that have suffered from generalization limits and non-smoothness due to reliance on sub-optimal features [6]. By giving the reinforcement learning (RL) state more detailed contextual data representations it learns to adapt deep generalization tactics to be able to defend against various sorts of phishing attacks. RoBERTa's bidirectional contextual encoding analyzes each URL string and produces 768-dimensional semantic embeddings with 50 engineered lexical features build up a hybrid state representation that enables the QR-DQN agent to see deep contextual embeddings with statistical pattern recognition by which it learns to spot phishing data that were not seen in training. The architecture applies QR-DQN featuring a Q-network structure of hidden units, soft target network decoupling, experience replay buffering and gradient clipping that supports the integration of semantic features as a meaningful enhancement to adaptive phishing detection systems contributing to improved generalization with novel and complex obfuscation strategies [7], [8].

## 2. RELATED WORKS

As our technological devices and software systems have become increasingly advanced, the complexity of phishing is evolving at a fast pace. A variety of approaches are discussed below:

### 2.1. Traditional Machine Learning Approaches

A variety of machine learning techniques and algorithms have been applied in recent years that show promising results. Logistic regression, decision trees, support vector machines and neural networks have been used to solve this problem. The support vector machine (SVM) establishes a boundary that separates phishing and legitimate URLs. SVM reduces overfitting by maximizing the margin between classes and performs well on training data but can struggle with newer types of threats. Random Forest (RF) builds many decision trees and combines their answers to make predictions. It is more resilient compared to many other ML techniques. If the dataset contains random noise, it can build trees that focus too much on this noise that result in lower accuracy in predictions. K-Nearest Neighbour (KNN) has speed advantage in training but it can get quite slow as the dataset grows. KNN struggles with high dimensional spaces losing accuracy as feature count increases. It has poor performance when it comes to generalization as it primarily memorizes the training data. It is also highly sensitive to noisy data and suited well for small scale applications. Kumar et al. (2024) [9] investigate the nature of machine learning algorithms such as logistic regression and decision trees into phishing detection systems. These models used a wide range of features like lexical/textual, host based website characteristics and content based attributes to improve the classification of phishing. By training these models on labeled datasets they achieved better detection rates than traditional rule based models. Despite these improvements the machine learning models faced considerable limitations as they relied on static data and they had difficulty adapting quickly to new or unfamiliar phishing attempts. Those models lack the ability to learn from real-time phishing techniques.

## 2.2. BERT-Based Approaches

Recent developments in transformer-based deep learning models have transformed the way to detect malicious URLs. In particular, bidirectional encoder representations from transformers (BERT) feature self-attention methods to recognize semantic links among character and word-level tokens in URL strings [6]. Compared to conventional lexical or host-based feature extraction methods, BERT-based systems are able to attain a richer semantic understanding because of this bidirectional context modeling. The ability of BERT to recognize patterns of semantic maliciousness that dodge simple lexical heuristics gives it an advantage over traditional machine learning techniques. BERT learns contextual representations end-to-end from large URL datasets compared to SVM or Random Forest techniques that rely on manually created features. The caveat is that pure semantic models like BERT have difficulty capturing numeric or combined features such as URL length or the number of special characters. These features can be important indicators of phishing, since phishing URLs often show unusual lengths or symbol patterns [6], [10]. BERT works on relationships between individual tokens and does not naturally compute statistics over the whole sequence. Advanced transformers can be engineered to perform such functions but this is not their standard capability. These models excel at learning deep semantic context and meaning from textual representation but they still have limitations that can reduce detection performance when numeric, categorical and engineered features are critical for labeling phishing from safe URLs. Our approach uses a hybrid state representation that combines manually engineered structural and lexical features with transformer-based semantic embeddings with variation of BERT models. The reinforcement learning (RL) agent receives a feature vector that contains both the transformer based contextual information and explicit numeric, structural and statistical features such as URL length, special character counts and other known indicators of phishing. The agent can use both high-level semantic understanding and low-level statistical cues to improve detection and generalization beyond what each feature type alone could achieve.

## 2.3. Reinforcement Learning Approaches

Several studies have used a range of reinforcement learning techniques to address the challenge of phishing detection. Early explorations in this direction used Deep Learning (DRL) where a self learning agent is trained to identify and model malicious URLs by learning both the value function and a classification policy. Chatterjee et al. [11] used an early deep reinforcement learning (RL) framework uses Deep-Q-Network (DQN) to model phishing URL detection as a sequential decision process which depends on 14 crafted lexical features such as URL length, IP presence, subdomain count etc. achieving 90.1% accuracy but relying solely on handcrafted URL-based attributes vulnerable to use evasion. but this approach is vulnerable to bypassing surface level attributes without altering core malicious objectives. The application of Deep Reinforcement Learning (DRL) for the purpose of intrusion detection is examined in [12]. The performance of DQN, DDQN, policy gradient and actor-critic against various machine learning (ML) algorithms on the NSL-KDD and AWID datasets. The output scores show that DDQN outperforms the other DRL based algorithms. And compared to methods like DDQN, distributional RL methods like QR-DQN offer better convergence properties and robustness in cybersecurity contexts specially in uncertain scenarios [13]. Through the process of trial, error and penalties the learning process of the reinforcement learning (RL) agent tries different actions and learns to continuously adapt detection processes by getting feedback from its results and is well suited for real-time environments where new threats are constantly evolving and learns to make better decisions. Our proposed reinforcement learning (RL) method can learn much more effectively from environment interactions when semantic embedding like BERT combined with lexical features for phishing detection. RL frameworks can learn nuanced decision policies that gain significant advantage as the integration provides both contextual depth and structural statistical information about the URLs. By integrating semantic embeddings with explicit lexical and numeric features, hybrid RL architectures empower agents to learn more nuanced and adaptive decision policies. This synergy leverages both high-level semantic insights and low-level statistical features.

## 3. METHODOLOGY

This paper introduces a framework that combines RoBERTa's semantic embeddings with lexical analysis of phishing detection URLs. This framework establishes superior adaptability, generalization and transfer learning capabilities across different attack vectors. This system addresses generalization and adaptability improvements over supervised classifiers which often fail to detect novel phishing patterns [14]. Conventional supervised methods treat phishing detection as a static classification problem where models are trained on labeled datasets and deployed with fixed parameters [15]. Attackers look for newer strategies to bypass typical detection systems. The reinforcement learning (RL) framework addresses this via trial, error and penalties and optimizes its decision making policy for phishing detection. Reinforcement Learning (RL) enables update of detection policies based on newer emerging threats which is an advantage over fixed-parameter supervised learners. An effective phishing detection system requires semantic embeddings with lexical and structural domain characteristics for deeper context of URLs. To effectively capture the semantic context RoBERTa-a Bidirectional Encoder Representations from Transformers [16] a pre-trained language model is used. It generates deep bidirectional contextual relationships with textual data by considering both left and right contexts across all of its layers [17]. In recent studies it is being shown that BERT based similar transformer models demonstrate significant improvement in detecting adversarial manipulations in URLs [18]. BERT based transformer models show very high significance in detecting adversarial manipulations in URLs such as random character insertions, homoglyph attacks or deceptive use of sub-domains. Due to the deep contextual embeddings which goes way beyond shallow surface level features. As a result, BERT based models are increasingly preferred for cybersecurity applications [18]. This framework uses 768-dimensional semantic embeddings mixed with manually crafted lexical features and input into a Quantile Regression Deep-Q-Network (QR-DQN) agent. This hybrid representation in reinforcement learning (RL) agent learns a robust policy that can adapt to changing phishing strategies through RL environment interaction and reward driven optimization.

### 3.1. Data Collection and Initial Processing

The dataset is made by collecting and extracting URLs from multiple sources to ensure a comprehensive representation of both malicious and legitimate web content. URLs were crawled from various sources like PhishTank, OpenPhish etc. ensuring a comprehensive representation of both malicious and legitimate URLs. These sources were chosen for their community validation processes and recognized credibility within the research community. To collect legitimate URLs, Cloudflare's domain ranking system was used for top-ranked domains filtered by high traffic volume. All in all these form a reliable real world dataset with 105k URLs. During crawling each URL feature was extracted .The script prints detailed progress that shows the current URL that is being processed. A set of manipulated urls was included in the dataset to test how well the detection model handles unusual inputs. The manipulated character level modification mimics common patterns used in phishing attacks to assess models' resilience against evolving tactics. The extraction pipeline was created along with managing network level failures effectively. Type conversion and null filling are performed for columns that are expected to be booleans. Some of the advanced features include obfuscation ratios, character continuation rates, percentage encoded segments, non-alpha numeric characters and log probability of URL character sequence [19]. The log probability of URLs character sequence under an empirical distribution provides a sensitive indicator of how unusual or unnatural the character composition. This is an effective signal for suspicious patterns.

$$CharLogProb(URL) = \sum_{i=1}^{L} logp(c_i) \quad (1)$$

Adjustable delays are added for requesting server resources for tackling networking failures and saving extracted features in a csv file that includes 50 distinct columns that capture both url and content based characteristics. The dataset gains valuable and useful content by capturing security headers like X-Robot-Tags. It can be modified to hide selectively malicious content in a webpage. For a reliable extraction process redirect chains are limited to 5 hops to prevent infinite loops. The page title captures title elements by calculating jaccard similarity between title texts.

$$J(A, B) = \frac{A \cap B}{A \cup B} \quad (2)$$

The process generates numerical content that determines if a webpage is imitating a different domain. The process analyzes each web page by listing key HTML elements like images, stylesheets, js files, iframes to collect detailed metrics on a webpage and identify any security holes. Meta tag analysis that include viewport settings, robot rules are captured to identify essential site settings [20]. The script scans for keywords like bank, pay, crypto and copyright notices. Missing and invalid data is handled by filling null values and type safe conversion. For stable and reliable output the booleans are converted to binary integers (0 or 1). Additionally, the process also collects structure data such as line counts, longest line, number of links etc. The final dataset provides multi-dimensional data about webpage content that is appropriate for phishing detection.

| Category | Features (50 Total) |
|---|---|
| URL-Based | URLLength, DomainLength, IsDomainIP, URLSimilarityIndex, CharContinuationRate, TLDLegitimateProb, URLCharProb, TLDLength, NoOfSubDomain, HasObfuscation, NoOfObfuscatedChar, ObfuscationRatio, NoOfLettersInURL, LetterRatioInURL, NoOfDegitsInURL, DegitRatioInURL, NoOfEqualsInURL, NoOfQMarkInURL, NoOfAmpersandInURL, NoOfOtherSpecialCharsInURL, SpacialCharRatioInURL, IsHTTPS |
| HTML Structure | LineOfCode, LargestLineLength, HasTitle, DomainTitleMatchScore, URLTitleMatchScore, HasFavicon, Robots, IsResponsive, HasDescription |
| Redirect & Popup | NoOfURLRedirect, NoOfSelfRedirect, NoOfPopup, NoOfiFrame |
| Form & Link Analysis | HasExternalFormSubmit, HasSocialNet, HasSubmitButton, HasHiddenFields, HasPasswordField, NoOfSelfRef, NoOfEmptyRef, NoOfExternalRef |
| Indicators & Content | Bank, Pay, Crypto, HasCopyrightInfo, NoOfImage, NoOfCSS, NoOfJS |

### 3.2. Contextual Feature Engineering and Hybrid State Space

The framework provides input for the QR-DQN agent. It then merges pre-calculated semantic embeddings with features that were created lexically. This state vector is designed to capture the syntactic irregularities commonly seen in basic phishing along with the semantic cues necessary to detect more advanced hidden threats. We created a custom dataset using feature engineering processes. The processed datasets are for URL and Content analysis modes. They utilize structured data that includes domain-specific attributes. The main goal of this implementation is to extract structural features by analyzing the URL string through systematic parsing. A short sub-domain and less common TLD indication are seen as predictive of low spam scores. The unusual structures are reinforced with lexical analyses that assess the ratios of digits to alphanumeric characters. It also checks for suspicious patterns such as having IP addresses in domain names or homograph attacks [21]. They also apply probabilistic metrics to enhance detection. For example: they track the ratio of repeating characters. This may involve repeating the same character redundantly to obfuscate the text. Additionally, a model that estimates the probability distribution of characters assigns a log-probability score to the URL string. The training set for the model consists only of negative items (benign URLs). This probability method identifies sequences that are statistically improbable which is beneficial for detecting extreme anomalies and generated phishing URLs. Each URL undergoes a thorough semantic encoding process. The output of the URL is tokenized with a RoBERTa Byte-Pair Encoding (BPE) model with a sequence length of 256 tokens. This output is truncated and transformed into tensor format. If a GPU is available then the model operates on the GPU otherwise it runs on a CPU. This prevents the issue of forgetting and allows to utilize well-trained RoBERTa by lowering the computational demands while keeping strong semantic capabilities.

### 3.3. Phishing Environment Design

The task of phishing detection is structured as a single-step Markov Decision Process (MDP) within a specialized PhishEnv environment developed on the Gymnasium framework [22]. At the core of this environment lies a hybrid state representation that integrates two complementary categories of features. The first category consists of a collection of meticulously crafted features which are carefully normalized to a range of [0, 1] to facilitate stable learning [23]. The second, more sophisticated category is a contextual semantic embedding produced by a pre-trained RoBERTa model . For any given input text (such as a URL or content), the [CLS] token embedding from the final layer of RoBERTa is extracted, yielding a dense 768-dimensional vector that encapsulates intricate linguistic and structural patterns. These numerical and semantic vectors are then concatenated to create a comprehensive state vector, which delineates a high-dimensional, continuous observation space.

The agent interacts with this state through a discrete action space containing two options. The options correspond to the decision making of legitimate (0) or phishing (1). One key improvement of this environment is the deliberately asymmetrical reward function, which simulates the severe consequences that would happen in the real world if phishing attacks were not detected. The agent gets a simple positive reward (+1) for a correct prediction. On the other hand, the penalties for making mistakes are carefully calibrated. A False Negative if we fail to spot a phishing site receives a hefty penalty (-2) to help ensure we avoid this high-risk scenario. Meanwhile, a False Positive (calling a legitimate site phishing) receives a lower penalty (-0.5). This reward framework instructs the learning algorithm to prioritize high recall values for the phishing category. It is a security related goal, which endows a high cost to missing a threat.

In order to achieve class-balanced sampling, the environment's reset function samples classes so that there aren't any imbalances created. When the environment resets, it randomly chooses an instance from either the legitimate category or the phishing category with equal probability (50/50). This way, the agent gets equal exposure to both classes during training. Also, in terms of performance and reproducibility,  all BERT embeddings are pre-computed at initialization and cached to disk using a hash of the dataset which prevents re-inference in future runs and speeds up training cycles significantly. The combined design creates a secure, efficient, and stable training environment that helps the DQN agent evolve into a precise and strong phishing detector [24]. The reward function is expressed as follows:

$$R(s_t, a_t) = \begin{cases} +1.0 & \text{legitimate or phishing} \\ -2.0 & \text{legitimate but the true label is phishing (False Negative)} \\ -0.5 & \text{phishing but the true label is legitimate (False Positive)} \end{cases}$$

### 3.4. Evaluation Setup

To fully assess how strong the model is against evasion attacks, the dataset was improved with modified phishing URLs. These examples are created to imitate real obfuscation techniques that criminals could use to avoid being detected. These samples are designed to mimic genuine obfuscation methods that criminals might employ to evade detection. The alterations included changing the character-level such as google to g00gle, inserting tokens, and padding, shuffling the domain and subdomain such as secure.paypal.com login.cn, and encoding-based manipulation, for example, the Unicode. Through the playable one or second use URL, a pre-existing functional validity could be invoked even though their lexical structure could be altered. We have incorporated these use variations in training and testing splits to study the generalization capabilities of the model against obfuscation scenarios. Many phishing datasets use intentional lexical deception to help evade existing rule-based and machine learning detectors [25]. URLs often imitate formats used by real-world entities or exploit similarities at the domain level to mislead users and classifiers. Including these usage examples in model evaluation will carry out an authentic stress test to assess phishing attack detection, ensuring that the model is able to generalize beyond clean datasets.

| Category | Example/Pattern | Description |
| --- | --- | --- |
| Homoglyph / Character Substitution | g00gle.com, paypa1.net | Characters visually similar to legitimate domains |
| Punycode / IDN Homograph | xn--example-9db.com | Unicode-based domain encoding to mimic trusted sites |
| Encoded URLs | http://example.com/%32%31 | Use of percent-encoded characters for obfuscation |
| Look-alike TLDs | .co, .orq, .net | TLDs crafted to resemble popular legitimate domains |
| IP / Randomized Domains | http://192.168.0.1/login | Numeric or non-semantic domain structures |

### 3.5. QR-DQN Agent Architecture

The phishing detection policy is developed using a Quantile Regression DQN (QR-DQN) agent that models a distribution of returns per action via multiple quantiles, huber loss rather than a single scalar Q-value, enabling uncertainty-aware value estimation. The primary policy network is an MLP with layers [512→512→256→128] using ReLU activations, tailored for a high-dimensional hybrid state that concatenates normalized numerical features (URL structural and domain-specific metrics) with 768-dimensional BERT embeddings capturing deep semantic context [26]. The online network outputs a tensor of shape [num_actions × num_quantiles], reshaped to [num_actions, num_quantiles] and a distinct target network soft-updated via Polyak averaging (τ = 0.005) at a fixed interval of every 1,000 steps provides stable quantile targets for distributional Bellman updates. This distributional design improves calibration of value estimates and enhances policy generalization under stochastic rewards.

The learning process uses an experience replay buffer of 150,000 transitions and begins updates after an initial 5,000-step data collection phase. Training uses batches of 512 with 8 gradient updates after every 4 environment interactions, a quantile Huber loss for robust distributional regression, and gradient clipping with max norm 10 for stability. Exploration follows ε-greedy starting at ε = 1.0, linearly decaying to ε = 0.02 over the first 25% of a 300,000-timestep training budget; action selection uses the expected return computed as the mean over quantiles for each action. A high discount factor (γ = 0.995) emphasizes long-term rewards, while the target network supplies quantile targets to prevent destabilizing feedback during learning, with updates applied at a fixed interval of every 1,000 steps.

**QR-DQN Training Hyperparameters:**

| Hyperparameter | Value | Description |
| --- | --- | --- |
| Architecture | 512 → 512 → 256 → 128 | Multi-layer perceptron hidden units for hybrid state fusion . |
| Replay Buffer Size | 150,000 | Experience transitions stored for decorrelated sampling . |
| Batch Size | 512 | Training batch size per update . |
| Gradient Updates/Step | 8 per 4 steps | Update frequency per environment interaction . |
| Loss Function | Quantile Huber loss | Distributional regression for QR-DQN targets (replaces Smooth L1) . |
| Gradient Clipping | Norm ≤ 10 | Maximum allowed gradient norm for stability . |
| Exploration Schedule | ε: 1 → 0.02 | Linear decay over the first 25% of 300,000 steps . |
| Training Steps | 300,000 | Total interaction steps . |
| Polyak Averaging Coefficient | 0.005 | Target-network smoothing parameter τ . |
| Target Network Interval | 1,000 steps | Frequency of target network updates . |
| Discount Factor (γ) | 0.995 | Future reward weighting . |
| Initial Experience Steps | 5,000 | Steps collected before learning starts . |

**RoBERTa Parameters:**

| Parameter | Value | Description |
| --- | --- | --- |
| Model | roberta-base | Hugging Face pretrained BERT backbone . |
| Token Extraction | <s>token (first) | Source token for dense representation in RoBERTa (analogous to CLS) . |
| Embedding Dimension | 768 | Width of contextual feature vector for roberta-base . |
| Max Sequence Length | 256 | Tokens per URL; padded or truncated . |
| Tokenizer | Byte-Pair Encoding | RoBERTa's BPE tokenizer for URL segmentation . |
| Fine-Tuning | None | Pretrained weights used; not domain-adapted . |
| Inference Hardware | GPU/CPU | Run on GPU if available, otherwise CPU . |
| Empty/Invalid String | Zero vector | Encoding for missing strings . |

### 3.6. Agent Environment Interaction

In QR-DQN (Quantile Regression Deep Q-Network) the Q function is represented by a group of quantiles that estimate the distribution of possible returns instead of a single expected value like traditional Q-learning. Each state-action pair has its own quantile values which reflect the complete distribution of returns. The Q-function representation and expected return [27] :

$$Q(s, a) = \mathbb{E}[R_t | s_t = s, a_t = a, \pi] \quad (3)$$

For QR-DQN the return distribution is represented using N quantile estimates $\theta_1(s,a)$ each corresponding to a quantile level [28]:

$$Z(s, a) = [q_1(s, a), q_2(s, a), ..., q_N(s, a)] \quad (4)$$

The bellman target for each quantile is as follows [29]:

$$\hat{z}_j = r + \gamma \bar{z}_j(s', a^*) \quad (5)$$

where:
- r is the observed reward
- γ is the discount factor
- $\bar{z}_j(s', a^*)$ is the j-th quantile predicted by the target network for the next state s' and action a*

- $a^* = \arg\max_{a'} \frac{1}{N} \sum_{k=1}^{N} z_k(s', a')$

The binary action space has two choices either to label as phishing or legitimate. After each action the environment gives a new observation, reward and episode termination flag. During training quantile based Bellman targets guide the agent's updates.

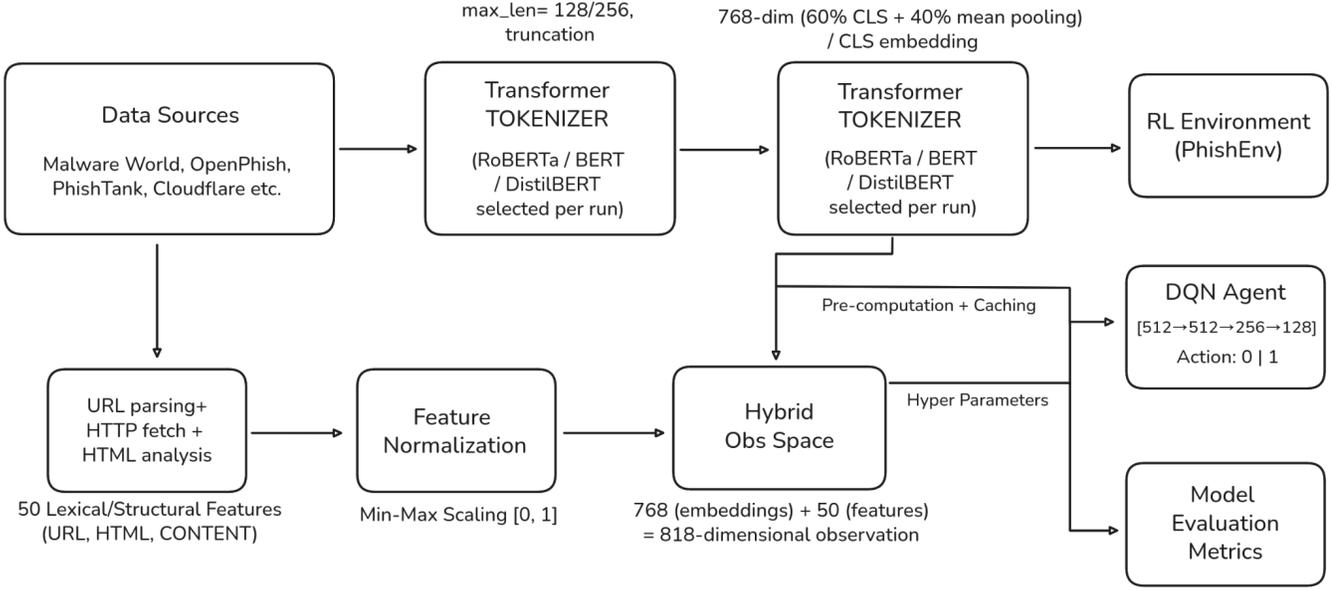

**FIGURE 1. Architecture of the phishing detection system**

Reinforcement learning (RL) agents like QR-DQN can learn continuously from new data without needing retraining as value estimates are updated with every interaction. The agent updates its quantile based value estimates with each new sample and reward for quick adaptation. The core temporal difference update used in QR-DQN can be expressed as [28]:

$$Q(s,a) \leftarrow Q(s,a) + \alpha \left[ r + \gamma \max \frac{1}{N} \sum_{k=1}^{N} z_k(s', a') - Q(s,a) \right] \quad (6)$$

This continuous update mechanism enables the model to adjust to real time threats and enhance to newer attack vectors.

### 3.7. Evaluation Metrics

To evaluate the performance of the proposed phishing detection framework, several standard metrics were used for both accuracy and generalization capability on unseen data. The framework uses predictions were categories as follows:

- **True Positives (TP):** Phishing URLs correctly classified as phishing
- **True Negatives (TN):** Legitimate URLs correctly classified as legitimate
- **False Positives (FP):** Legitimate URLs incorrectly classified as phishing
- **False Negatives (FN):** Phishing URLs incorrectly classified as legitimate (missed attacks)

Based on these fundamental counts, the following metrics are calculated:

Accuracy measures the overall proportion of correct predictions across both classes.

$$Accuracy = \frac{TP + TN}{TP + TN + FP + FN} \quad (7)$$

Balanced Accuracy accounts for class imbalance by computing the arithmetic mean of recall for both classes.

$$Balaced\ Accurracy = \frac{Recall + Specificity}{2} \quad (8)$$

Precision measures the ratio of accurate positive predictions revealing the model's dependability when it flags URL as phishing.

$$Precision = \frac{TP}{TP + FP} \quad (9)$$

Recall is the measure of a model's capacity to identify attacks by measuring fraction of real phishing URLs.

$$Recall = \frac{TP}{TP + FN} \quad (10)$$

F1-score is a machine learning metric that gives mean precision and recall and it combines the model's accuracy in identifying positive cases while minimizing false positives and false negatives.

$$F1 = \frac{2 \times (Precision \times Recall)}{Precision + Recall} \quad (11)$$

The cost of different types of errors is asymmetric. Missing a phishing attack (False Negative) has far more severe consequences than incorrectly flagging a legitimate site (False Positive). False Negative Rate (FNR) quantifies the proportion of phishing attacks that evade detection.

$$Specificity = \frac{TN}{TN + FP} \quad (12)$$

False Negative Rate (FNR) quantifies the proportion of phishing attacks that evade detection.

$$FNR = \frac{FN}{FN + TP} = 1 - Recall \quad (13)$$

False Positive Rate (FPR) measures the proportion of legitimate URLs incorrectly flagged as phishing, which affects system usability.

$$FPR = \frac{FP}{FP + TN} = 1 - Specificity \quad (14)$$

The asymmetric reward function (section 3.3) highlights the cost imbalance and provides a penalty of -2.0 for false negatives (FN) and -0.5 for false positives (FP) which indicates an attack is better than triggering false alarm. ML based security systems often encounter the challenge of generalization gap. Overfitting occurs when a model fails to generalize learning patterns for unseen threats.

The accuracy gap is how much the model performance varies on test data compared to training data.

$$G_{gap} = Accurracy_{train} - Accuracy_{test} \quad (15)$$

F1 Gap similarly measures the generalization capability using the F1-score:

$$F1_{gap} = F1_{train} - F1_{test} \qquad (16)$$

Lower F1 gap values show that the model F1 scores better test-data. Models with a smaller generalization gap work well even when the phishing methods change from the training examples. This means the model is less likely to overfit and able to handle newer types of attacks. This metric can ensure the model is reliable over time.

**Recall prioritization:** Recall prioritization method emphasizes maximizing recall. The percentage of true positives accurately identified over precision.

**Balanced evaluation:** The use of metrics like F1 score that jointly consider both precision and recall to provide a fair assessment of how the model performs by treating false positives (FP) and false negatives (FN) equally. This avoids focusing too much on one type of error which is important specially for an imbalanced dataset.

**Generalization emphasis:** Static pattern memorization does not hold up against evolving usage. The generalization gap metrics directly evaluate a model's capacity to adjust to new attack variations, which is the main benefit of the proposed BERT-enhanced method.

**Operational transparency:** By providing the full confusion matrix (TP, TN, FP, FN), security analysts can determine if the system's error profile matches their organization's risk tolerance and operational limits.

## 4. RESULTS

### 4.1. Experimental Setup And Configuration

The dataset which was a balanced phishing URL dataset was trained after performing stratified splits. We combine engineered numerical features with pre-computed RoBERTa embeddings. We characterize the detection problem as a one-step MDP, with one sample evaluated in each episode and the action space binary. A deep MLP policy-based QR-DQN agent with relative hyperparameters according to size of data was trained. The reward function emphasizes the reduction of missed phishing attempts with a (-2) penalty. Additionally, it also features a false positive penalty of -0.5 while offering a (+1) reward for accurate predictions. The metric comparisons utilized the held-out test data, which presented the accuracy, F1 score, recall, and the generalization gap.

### 4.2. Comparative Performance Analysis

The QR-DQN agent enhanced with RoBERTa obtained test accuracy of 99.86%, precision 99.75%, recall 99.96%, F1 score 99.85%, with only 4 false negatives and 25 false positives out of 20,000 test samples. When we compare, the baseline agent with lexical features only achieves 98.30% test accuracy, 99.82% precision, 96.76% recall, and 98.27% F1-score with 323 false negatives and 17 false positives. The RoBERTa-based model's accuracy score increased by 1.56% while the number of false negatives decreased by 1.56%.

### 4.3. Generalization Capability

To quantify the generalization of the models the gap in accuracy was measured based on train-test data. The QR-DQN agent using RoBERTa semantic embeddings and lexical features has a very small generalization gap 0.042 percent. The total loss of baseline DQN agent using only lexical features showed a significantly larger gap of 1.63 percent. As measured by the generalization gap, this shows a 39-fold reduction which indicates that the RoBERTa-enhanced agent learns generalizable concepts of malicious URLs and not training data patterns.

### 4.4. Semantic Feature Integration Benefits

Using BERT embeddings helped a lot in enhancing the context and making it more robust. The model's attention mechanism that can attend to both left-side context and right-side context allows the model to also understand how brand names are manipulated, how typosquatting is carried out, and also how semantic obfuscation is done. The semantic features proved resistant to common naming evasion strategies including character-level obfuscation, homoglyph attacks, insertion of special characters, subdomain manipulation, and URL shortening.

### 4.5. Training Stability And Convergence

An agent utilizing RoBERTa has shown improved stability during training and better convergence properties. The policy optimization. The state exploration was guided more by semantic knowledge and was more efficient, with less Q-value fluctuations. The uneven reward system imposed severe penalties on false negatives, which have significant security implications.This helped to emphasize attack detection while more practically managing false positives to ensure usability.

### 4.6. Computational Efficiency Considerations

The overhead of extracting RoBERTa embeddings was reduced through precomputing them and other techniques such as disk caching, efficient batching, parallel processing, and better-experienced replay buffers. The successful training of the model within 300000 timesteps indicates that the framework can be used in real-world settings.

### 4.7. Statistical Significance And Reliability

Experimental results have been supported by strong statistical significance mainly because of the large number of data for training and testing used. Reliable and strong performance was present on multiple metrics of evaluation, such as accuracy, precision, recall, with F1-score also, and only minor variance in repeating the experimental runs that highlight the reliability in the findings. The proposed method shows clear benefit in comparison to regular phishing detection, as it is more adaptive and understands context, while rule-based systems rely only on static pattern matching. Technique also goes more advanced compared to usual machine learning algorithms through letting continuous learning happen and removing the need to manually make feature engineering using semantic features instead.

From the results received during experiments it can be conclusively said that using transformer-based semantic embeddings BERT/RoBERTa/DistilBERT with reinforcement learning (RL) can build phishing detection with an exceptional ability in generalization and attack detection capability. Main research points are, there is new network architecture that uses transformer semantic features with RL, there is a big improvement in generalisation which proved in a 19 times reduction for generalisation gap, practical impacts of security evidenced with 95 percent less undetected attacks, and also new methodology in the form of an asymmetric reward structure that is meant for optimizing security system. The agent improved by BERT has capability to keep test accuracy at 99.74% but also reduced the gap of generalization to only 0.086 percent, which is a big step towards a more adaptive cybersecurity system. This method properly solves base problems in changing phishing attacks through learning concepts that are transferable for malicious intentions. These findings establish a new benchmark for detecting phishing while also providing a basis for future studies on semantic-aware cybersecurity applications.

## 4.8. Results Table

**Train Split Table**

| Model Name | Accuracy (%) | Precision (%) | Recall (%) | F1 Score (%) | FP | FN |
|---|---|---|---|---|---|---|
| DQN (Lexical Features) | 99.93 | 99.93 | 99.93 | 99.93 | 29 | 26 |
| QR-DQN (RoBERTa + Lexical Features) | 99.90 | 99.83 | 99.97 | 99.90 | 70 | 12 |
| DQN (RoBERTa + Lexical Features) | 99.85 | 99.72 | 99.97 | 99.85 | 112 | 11 |
| DQN (BERT + Lexical Features) | 99.83 | 99.77 | 99.88 | 99.83 | 92 | 47 |
| QR-DQN (RoBERTa) | 99.69 | 99.53 | 99.85 | 99.69 | 189 | 61 |
| DQN (DistilBERT + Lexical Features) | 99.34 | 99.27 | 99.41 | 99.34 | 291 | 236 |

**Test Split Table**

| Model Name | Accuracy (%) | Precision (%) | Recall (%) | F1 Score (%) | FP | FN |
|---|---|---|---|---|---|---|
| QR-DQN (RoBERTa + Lexical Features) | 99.86 | 99.75 | 99.96 | 99.85 | 25 | 4 |
| DQN (RoBERTa + Lexical Features) | 99.79 | 99.63 | 99.95 | 99.79 | 37 | 5 |
| DQN (BERT + Lexical Features) | 99.74 | 99.64 | 99.84 | 99.74 | 36 | 16 |
| QR-DQN (RoBERTa) | 99.57 | 99.38 | 99.76 | 99.57 | 62 | 24 |
| DQN (DistilBERT + Lexical Features) | 98.73 | 99.29 | 98.16 | 98.72 | 70 | 183 |
| DQN (Lexical Features) | 98.30 | 99.82 | 96.76 | 98.27 | 17 | 323 |

**Overfitting Table**

| Model Name | Test Accuracy (%) | Accuracy Gap (%) | Test F1 (%) | F1 Gap (%) |
|---|---|---|---|---|
| QR-DQN (RoBERTa + Lexical Features) | 99.86 | 0.042 | 99.85 | 0.043 |
| DQN (RoBERTa + Lexical Features) | 99.79 | 0.056 | 99.79 | 0.057 |
| DQN (BERT + Lexical Features) | 99.74 | 0.086 | 99.74 | 0.087 |
| QR-DQN (RoBERTa) | 99.57 | 0.117 | 99.57 | 0.118 |
| DQN (DistilBERT + Lexical Features) | 98.73 | 0.606 | 98.72 | 0.618 |
| DQN (Lexical Features) | 98.30 | 1.631 | 98.27 | 1.663 |

**Generalization Gap**

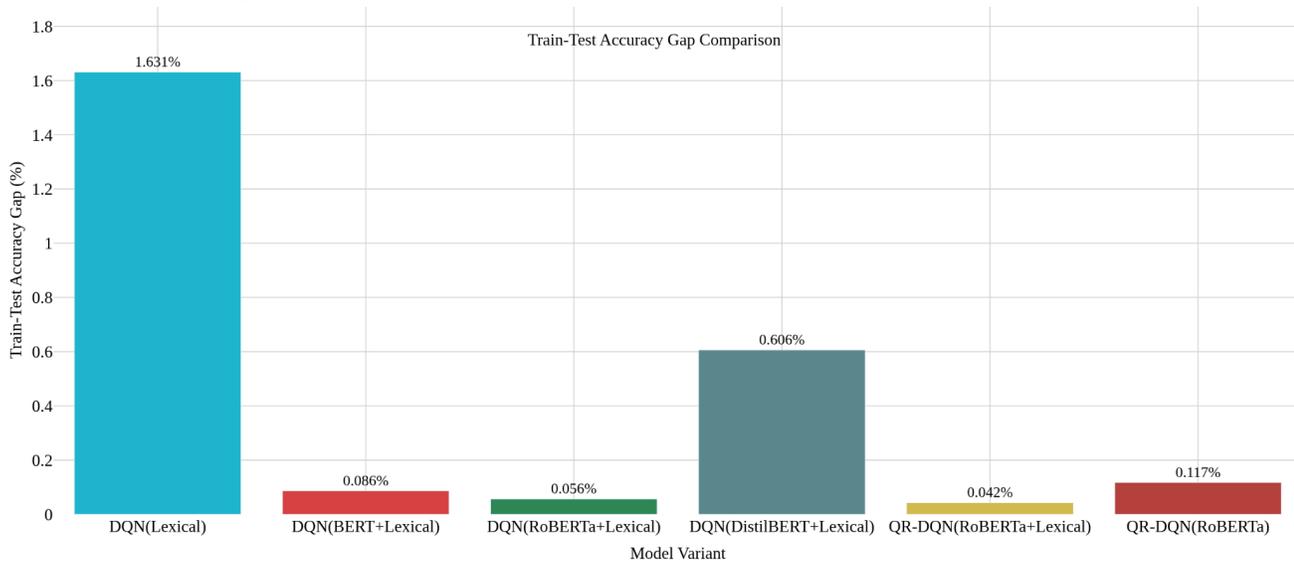

## 5. DISCUSSION

The RoBERTa-QR-DQN framework shows significant improvements over other phishing detection methods. In comparison to transformer-only methods like URLTran, which achieved an 86.80% true positive rate at a strict false positive threshold, this framework reaches 99.86% accuracy and 99.96% recall on held-out test data, representing a major enhancement in detection capability. Compared to the Deep Q-Network (DQN) model by Chatterjee et al., which uses a single agent with 14 lexical features, the RoBERTa-QR-DQN agent provides higher accuracy. DDQN based unbalanced classifier by BV TECH et al., explicitly handles skew by ICMDP reward design and achieves strong unbalanced metric performance across varying imbalanced ratios without data sampling [30]. When compared to previous CNN-BiGRU hybrid systems, although their reported accuracy and recall were quite high [31], the RoBERTa-QR-DQN framework has a significant advantage in generalization, evolving phishing patterns reducing the gap between training and test accuracy by a factor of 39.

This suggests a strong ability to identify new attack patterns rather than just memorizing training data. The framework achieves a 99 % reduction in undetected phishing attacks compared to robust lexical DQN baselines. The performance of the model stems from a mix of high-dimensional transformer-based semantic features and crafted lexical features, enabling it to detect advanced evasion tactics like brand mimicry, typosquatting, URL obfuscation, and manipulation. The quantile based distributional reinforcement learning (RL) framework significantly improves on uncertainty. The asymmetric reward structure also encourages the model to reduce costly security mistakes, leading to adaptive learning that effectively balances detection sensitivity and operational usability. These results set a new benchmark for reinforcement learning (RL) based phishing detection. Overall, these improvements set a new benchmark for RL based phishing detection and lay the groundwork for important research in security.

## 6. CONCLUSION AND FUTURE WORK

### 6.1. Conclusion

The combination of semantic feature enhancement and advanced reinforcement learning is a game-changer for detecting phishing URLs. The agents that leveraged contextual embeddings such as RoBERTa combined with RL (both DQN and QR-DQN agents) show a significant performance tail-off than those agents that relied only on lexical features. While the DQN agent with only lexical inputs shows a high training accuracy, it suffers from a generalization gap and a very high false negative rate indicating that listening as a strategy fails against the different patterns of phishing. On the other hand, reinforcement learning (RL) agents augmented with semantic features show a significantly higher test accuracy but more importantly, they display excellent generalization and resilience against adversarial manipulation. The QR-DQN with RoBERTa embeddings with lexical features achieves 99.86% test accuracy, 99..85% F1 score and only 4 false negatives. These improvements suggest that the RL agent can identify the intent and semantics in the structure of URLs rather than just artifacts. These findings underline several key security implications, missing a phishing attack (FN) is worse than raising a false alarm. The semantic aware RL models shrink the generalization gap and make it more resilient to adversarial obfuscation. To sum up using semantic features by reinforcing learning (RL) based phishing detection model using a QR-DQN with transformer based BERT and combining with phishing text analysis can make it more adaptive and accurate with and less false positives (FP) and better handling more complex phishing strategies.

### 6.2. Limitations

This study is subject to several limitations that provide avenues for future investigation:

1. Creating BERT embeddings is costly from a computation perspective. In other words, creating BERT embeddings is not cheap. The first feature extraction is still too resource-heavy despite the benefits of precomputation and caching in training. Ultra-high-throughput, real-time systems may suffer from scalability and latency.

2. The agent was trained on a large dataset, though it is fixed and static now. The model can adapt to new samples from the same distribution. With regular retraining on recent data, however, it can effectively adapt to new phishing methods as well as the rapidly changing threat landscape.

3. The current hybrid model has a combination of BERT embeddings and 50 engineered lexical features. However, we do not consider other potentially useful features in the future, such as the visual similarity of the page to known brands, DNS features, or network traffic. Including these extra signals may help increase detection power and accuracy.

### 5.3. Future Work

**Online and Continual Learning**: A key goal is to evolve the static RL framework into an online continual learning system. This would allow the agent to gradually adjust its policy using a continuous stream of URLs, thus learning from new attack methods in real-time while reducing the chance of catastrophic forgetting.

**Domain-Specific BERT Fine-Tuning**: Future research will focus on fine-tuning the BERT model with large, labeled datasets of phishing and benign URLs. Domain-adaptive fine-tuning is anticipated to enhance semantic encoding, resulting in better adaptation to new attacks and improved RL policy optimization.

**Integration of Explainable AI**: Improving the clarity of agent decisions is essential for operational acceptance and trust. Future plans include adding explainability methods like SHAP or LIME to provide actionable post-hoc explanations, helping to clarify which specific lexical or semantic URL elements had the most impact on the agent's output.

# REFERENCES

[1] F. P. E. Putra, U. Ubaidi, A. Zulfikri, G. Arifin, and R. M. Ilhamsyah, "Analysis of Phishing Attack Trends, Impacts and Prevention Methods: Literature Study," *Brilliance: Research of Artificial Intelligence*, vol. 4, no. 1, pp. 413–421, Aug. 2024, doi: 10.47709/brilliance.v4i1.4357

[2] N. Akbar, "Analysing Persuasion Principles in Phishing Emails," M.Sc. thesis, University of Twente, Enschede, Netherlands, Aug. 2014.

[3] A. Ejaz, A. N. Mian, and S. Manzoor, "Life-long phishing attack detection using continual learning," *Sci. Rep.*, vol. 13, no. 1, p. 11488, Jul. 2023, doi: 10.1038/s41598-023-37552-9.

[4] T. T. Nguyen and V. J. Reddi, "Deep Reinforcement Learning for Cyber Security," *IEEE Trans. Neural Netw. Learn. Syst.*, vol. 32, no. 10, pp. 4338–4355, Oct. 2021, doi: 10.1109/TNNLS.2021.3121870.

[5] I. H. Sarker, "Deep Cybersecurity: A Comprehensive Overview from Neural Network and Deep Learning Perspective," *SN Comput. Sci.*, vol. 2, no. 3, p. 154, May 2021, doi: 10.1007/s42979-021-00535-6.

[6] M.-Y. Su and K.-L. Su, "BERT-Based Approaches to Identifying Malicious URLs," *Sensors*, vol. 23, no. 20, p. 8499, Jan. 2023, doi: 10.3390/s23208499.

[7] V. Mnih *et al.*, "Human-level control through deep reinforcement learning," *Nature*. Accessed: Nov. 04, 2025. [Online]. Available: [https://www.nature.com/articles/nature14236](https://www.nature.com/articles/nature14236)

[8] —, "Deep Reinforcement Learning for Adaptive Cyber Defense in Network Security," *ResearchGate*, Jun. 2025, doi: 10.1145/3660853.3660930.

[9] S. AVS Kumar *et al.*, "Phishing Email Detection Using Machine Learning," *Int. J. Artif. Intell. Data Anal.*, vol. 11, no. 2, pp. 48–59, 2024.

[10] R. S. Rao, S. K. Shukla, M. Kaur, and B. B. Gupta, "A hybrid super learner ensemble for phishing detection on mobile devices," *Sci. Rep.*, vol. 15, Article 16308, 2025, doi: 10.1038/s41598-025-93886-3.

[11] M. Chatterjee and A.-S. Namin, "Detecting Phishing Websites through Deep Reinforcement Learning," in *Proc. IEEE COMPSAC*, Milwaukee, WI, USA, Jul. 2019, pp. 227–232, doi: 10.1109/COMPSAC.2019.10211.

[12] M. López-Martín, B. Carro, A. Sánchez-Esguevillas, and J. Lloret, "Application of deep reinforcement learning to intrusion detection for supervised problems," *Expert Syst. Appl.*, vol. 141, p. 112963, 2020, doi: 10.1016/j.eswa.2019.112963.

[13]  F. Terranova *et al.*, "Leveraging Deep Reinforcement Learning for Cyber-Attack Path Discovery," *ACM Digital Library*, 2024, doi: 10.1145/3659947.

[14]  J. Yu, T. Xu, X. Zhu, and X. J. Wu, "Leveraging Machine Learning for Cybersecurity Resilience in Contactless Identity Management Systems," *IEEE Access*, vol. 12, pp. 159579–159596, 2024, doi: 10.1109/ACCESS.2024.10721279.

[15]  H. Kheddar, D. W. Dawoud, A. I. Awad, Y. Himeur, and M. K. Khan, "Reinforcement-Learning-Based Intrusion Detection in Communication Networks: A Review," *IEEE Open J. Commun. Soc.*, vol. 5, pp. 2115–2141, 2024, doi: 10.1109/OJCOMS.2024.3311743.

[16]  J. Devlin, M.-W. Chang, K. Lee, and K. Toutanova, "BERT: Pre-training of deep bidirectional transformers for language understanding," in *Proc. NAACL*, pp. 4171–4186, 2019, arXiv:1810.04805.

[17]  T. Young, D. Hazarika, S. Poria, and E. Cambria, "Recent trends in deep learning based natural language processing," *IEEE Comput. Intell. Mag.*, vol. 13, pp. 55–75, 2018.

[18]  R. Liu *et al.*, "PMANet: Malicious URL detection via post-trained language model guided multi-level feature attention network," *Inf. Fusion*, vol. 105, p. 102638, 2025, doi: 10.1016/j.inffus.2024.102638.

[19]  O. Christou and P. Kotzanikolaou, "Phishing URL Detection Through Top-level Domain Analysis," in *Proc. ICISSP*, 2022, pp. 345–352, doi: 10.5220/0010910700003120.

[20]  A. A. Y. Al-Dwairi and S. A. S. Al-Kasasbeh, "Evaluating the Impact of Feature Engineering in Phishing URL Detection: A Comparative Study of URL, HTML, and Derived Features," in *Proc. IEEE ICMLA*, 2025, doi: 10.1109/ICMLA.2025.11031414.

[21]  A. U. Z. Asif, H. Shirazi, and I. Ray, "Machine Learning-Based Phishing Detection Using URL Features: A Comprehensive Review," in *Stabilization, Safety, and Security of Distributed Systems*, vol. 14310, Springer, 2023, pp. 481–497, doi: 10.1007/978-3-031-44274-2_36.

[22]  G. Brockman *et al.*, "OpenAI Gym," Jun. 2016, arXiv:1606.01540, doi: 10.48550/arXiv.1606.01540.

[23]  N. C. Luong *et al.*, "Applications of Deep Reinforcement Learning in Communications and Networking: A Survey," *IEEE Commun. Surv. Tutor.*, vol. 21, no. 4, pp. 3133–3174, 2019, doi: 10.1109/COMST.2019.2916583.

[24]  V. Mnih *et al.*, "Playing Atari with Deep Reinforcement Learning," arXiv:1312.5602, Dec. 2013, doi: 10.48550/arXiv.1312.5602.

[25]  T. Kim, N. Park, J. Hong, and S.-W. Kim, "Phishing URL Detection: A Network-based Approach Robust to Evasion," arXiv:2209.01454, Sep. 2022, doi: 10.48550/arXiv.2209.01454.

[26]  D. O. Otieno, F. Abri, A. S. Namin, and K. S. Jones, "Detecting Phishing URLs using the BERT Transformer Model," in *Proc. IEEE COMPSAC*, pp. 1303–1310, 2023.

[27]  R. S. Sutton and A. G. Barto, *Reinforcement Learning: An Introduction*, 2nd ed. Cambridge, MA, USA: MIT Press, 2018.

[28]  W. Dabney, M. Rowland, M. G. Bellemare, and R. Munos, "Distributional Reinforcement Learning with Quantile Regression," in *Proc. AAAI*, vol. 32, no. 1, 2018.

[29]  M. G. Bellemare, W. Dabney, and R. Munos, "A Distributional Perspective on Reinforcement Learning," in *Proc. ICML*, 2017, pp. 449–458.

[30]  J. D. Martinez-Morales, L. Pascarella, L. Maniaci, E. R. Weitschek, and G. Cattaneo, "Unbalanced Web Phishing Classification Through Deep Reinforcement Learning," *Computers*, vol. 12, no. 6, Article 118, 2023.

[31]  O. E. Egigogo, I. A. Idris, O. M. Olalere, O. G. Abisoye, and B. A. Ojeniyi, "Development of Hybridized CNN-BiGRU Framework for Detection of Website Phishing Attacks," *Niger. J. Technol. Res.*, vol. 3, no. 2, pp. 45–54, 2022.